\documentclass[letterpaper, 10 pt, conference]{ieeeconf}  

\IEEEoverridecommandlockouts                              

\usepackage{graphics} 
\usepackage{epsfig} 
\usepackage{mathptmx} 
\usepackage{times} 
\usepackage{amsmath} 
\usepackage{amssymb}  
\usepackage{graphicx}
\usepackage{capt-of}
\usepackage{balance}

\title{\LARGE \bf
ThorArena: Benchmarking Humanoid Physical Interaction with Human Motion-Force Demonstrations
}

\author{
Chenhao Yu$^{1,*}$,
Hongwu Wang$^{1,*}$,
Weitao Zhang$^{1,*}$, \\
Youhao Hu$^{1}$,
Jiachen Zhang$^{1}$,
Gangyang Li$^{1}$,
Alois Knoll$^{2}$,
Shaqi Luo$^{1,\dagger}$\\
$^{1}$Beijing Academy of Artificial Intelligence
$^{2}$Technical University of Munich
\thanks{$^{*}$These authors contributed equally.}
\thanks{$^{\dagger}$Corresponding author.}
}

\IEEEaftertitletext{%
\vspace{-1em}
\begin{center}
    \includegraphics[width=0.95\textwidth]{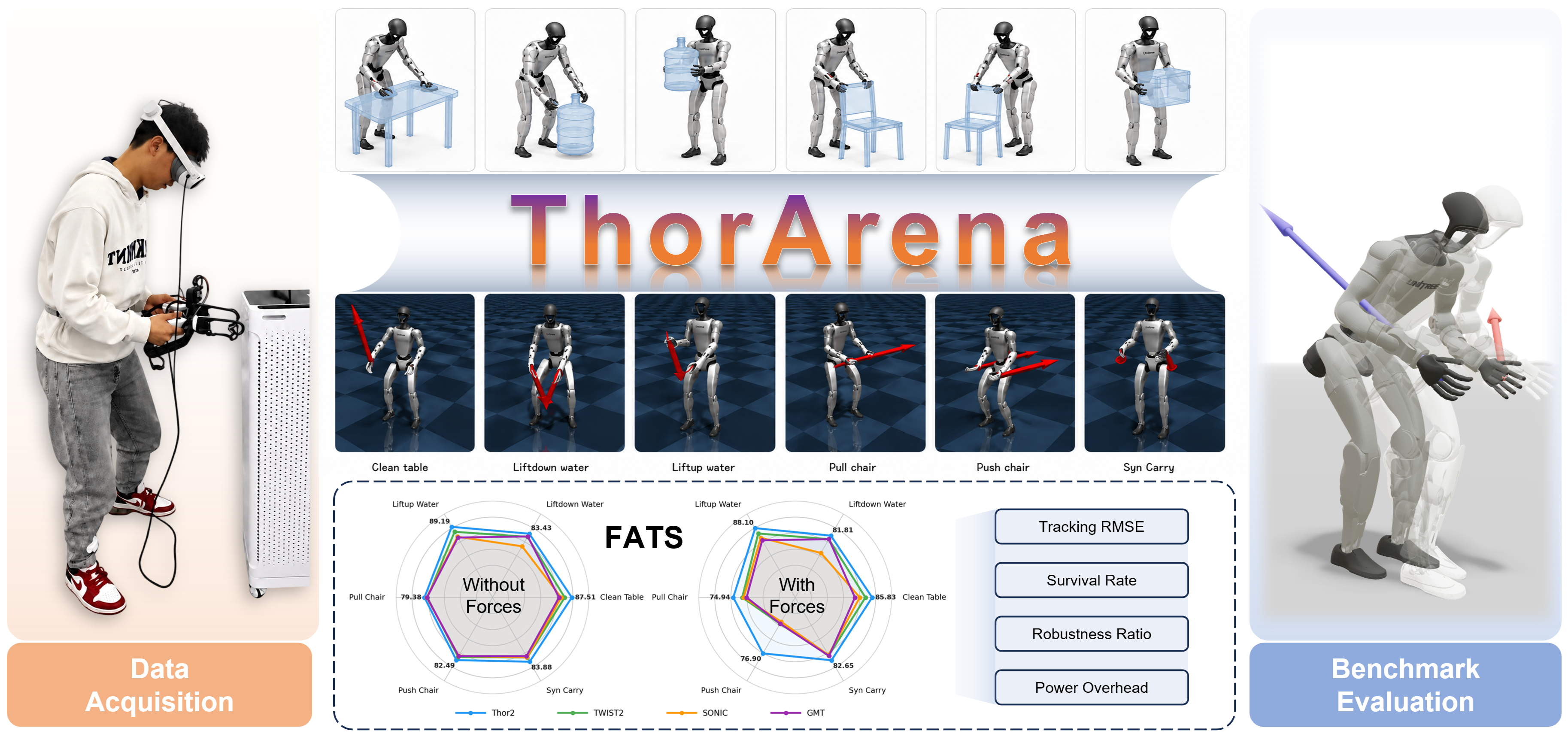}
    \captionof{figure}{
    \textbf{Overview of ThorArena.}
    ThorArena integrates real-world motion--force demonstration collection, synchronized interaction-force replay in simulation, and standardized evaluation of humanoid whole-body control policies.
    The acquisition system records synchronized whole-body human motion and forces exerted through both hands across six representative physical interaction tasks.
    The captured motions are retargeted to humanoid reference trajectories, while the corresponding recorded forces are synchronously replayed at the robot hands during simulation.
    Policies are evaluated under matched force and no-force settings using FATS and complementary diagnostic metrics, including tracking RMSE, survival rate, robustness ratio, and power overhead.
    }
    \label{fig:main}
\end{center}
\vspace{0.5em}
}

\begin{document}

\maketitle
\thispagestyle{empty}
\pagestyle{empty}


\begin{abstract}

Humanoid robots are increasingly expected to perform contact-rich tasks that require not only accurate whole-body motion but also robust physical interaction with surrounding objects and humans. Although recent advances in humanoid motion imitation and whole-body control have achieved remarkable tracking performance, existing datasets and benchmarks primarily focus on kinematic motion while largely overlooking synchronized interaction forces. As a result, current evaluations fail to capture how external interaction forces affect tracking accuracy, stability, and control robustness.

In this paper, we present ThorArena, a benchmark for evaluating force-aware humanoid interaction based on human demonstrations with synchronized motion and force measurements. We collect a real-world interaction dataset that simultaneously captures whole-body human motion and forces exerted by both hands across six representative physical interaction tasks. Based on these demonstrations, we propose force-aware evaluation metrics that jointly assess whole-body tracking accuracy, robustness under different force levels, control effort, and episode survival through the Force-Aware Tracking Score (FATS) and complementary diagnostic metrics. We further establish a unified benchmark protocol that replays recorded interaction forces in simulation and provides a standardized evaluation interface for different humanoid control policies.

Experiments on representative whole-body control policies demonstrate that force-aware evaluation reveals substantial performance differences that remain largely hidden under conventional no-force evaluation. ThorArena provides a practical and reproducible framework for studying force-aware humanoid interaction and offers a new benchmark for evaluating contact-rich humanoid behaviors.

\end{abstract}

\section{INTRODUCTION}

Humanoid robots are increasingly being developed for real-world applications, where many tasks require not only coordinated whole-body motion but also physical interaction with surrounding objects and humans. Recent advances in humanoid teleoperation, motion imitation, and whole-body control have enabled robots to reproduce increasingly complex human behaviors through large-scale human demonstrations and simulation-based policy learning. However, existing humanoid datasets and benchmarks primarily emphasize kinematic motion tracking, task completion, or motion naturalness, while the role of measured interaction forces is largely overlooked. As a result, current evaluation protocols provide only limited insight into how humanoid policies behave under realistic physical interactions.

Force interaction is a fundamental component of contact-rich humanoid
tasks. When a humanoid physically interacts with surrounding objects, the
resulting forces directly affect whole-body balance, motion tracking
accuracy, and control effort. A policy that performs well in free-space
motion tracking may experience substantial performance degradation or even
failure when interaction forces are applied to the upper body. Therefore,
evaluating humanoid behaviors solely under kinematic conditions is
insufficient for assessing their performance during physical interaction.
To properly characterize force-aware humanoid control, benchmark protocols
should evaluate not only motion tracking accuracy but also robustness to
external interaction forces and stability over long-horizon execution.

Although recent humanoid datasets have significantly expanded the diversity of motion demonstrations, most of them only provide pose trajectories or retargeted motions without synchronized interaction forces. Likewise, existing evaluation protocols mainly measure kinematic errors or task success while treating external interactions as implicit disturbances. Consequently, there remains a lack of standardized benchmarks that jointly incorporate human motion demonstrations, measured interaction forces, and force-aware evaluation metrics into a unified evaluation framework.

In this work, we present \textbf{ThorArena}, a benchmark for force-aware humanoid interaction using synchronized human motion and force demonstrations. ThorArena consists of three tightly integrated components. First, we collect a force-interaction humanoid dataset using a real-world acquisition system that simultaneously records human body motion and two-hand interaction forces, covering six representative contact-rich tasks. Second, we introduce force-aware evaluation metrics that jointly characterize whole-body tracking accuracy, robustness under different force levels, control effort, and episode survival through the proposed Force-Aware Tracking Score (FATS) together with complementary diagnostic metrics. Third, we establish a simulation-based benchmark that replays recorded interaction forces during policy rollout and provides a unified force-replay protocol and policy-adapter interface for evaluating different humanoid control policies under consistent physical disturbances.

By combining synchronized motion-force demonstrations with standardized evaluation protocols, ThorArena enables systematic assessment of humanoid policies in realistic contact-rich scenarios. Compared with conventional motion benchmarks, it explicitly measures how interaction forces influence tracking performance, robustness, and control behavior, providing a practical benchmark for future research on force-aware humanoid interaction.

The main contributions of this work are summarized as follows:

\begin{itemize}
    \item We collect and release a \textbf{force-interaction humanoid dataset}
    using a real-world acquisition system, containing synchronized human body
    motion and two-hand interaction-force demonstrations across six
    representative contact-rich tasks. The paired motion--force data have been
    used to train the force-aware humanoid control policy Thor2, illustrating
    their applicability to policy learning and providing a valuable resource
    for future research on force-aware humanoid control.

    \item We propose \textbf{force-aware evaluation metrics}, including the Force-Aware Tracking Score (FATS), which jointly evaluate whole-body tracking accuracy, robustness under different force levels, control effort, and episode survival.

    \item We present \textbf{ThorArena}, a simulation-based benchmark for force-aware humanoid interaction with a unified force-replay protocol and policy-adapter interface, enabling consistent evaluation of representative humanoid whole-body control policies under realistic interaction forces.
\end{itemize}

\section{RELATED WORK}

\subsection{Humanoid Teleoperation and Motion Retargeting}

Humanoid teleoperation transfers human motion and task commands to humanoid
robots and has become an important tool for whole-body control and
demonstration collection. Recent systems support real-time teleoperation
through vision-based pose estimation, VR devices, motion capture systems,
and exoskeleton interfaces
\cite{he2024learning,fu2024humanplus,he2024omnih2o,ben2025homie,ze2025twist,ze2025twist2}.
Closed-loop approaches further improve long-horizon execution by correcting
accumulated global pose drift
\cite{li2025clone,zhu2026clot}.

Motion retargeting bridges the embodiment gap between humans and humanoid robots. Differences in body proportions, joint limits, and actuation capabilities can introduce foot sliding, penetration, self-collision, and dynamically infeasible motions. Existing methods address these issues using
inverse kinematics, motion filtering, data augmentation, dynamics-aware optimization, and learned tracking policies \cite{Luo2023PerpetualHC,joao2025gmr,yang2025omniretarget}.
Recent work further improves tracking generality through unified command spaces, adaptive sampling, mixture-of-experts architectures, and large-scale
motion models \cite{he2025hover,li2025amo,chen2025gmt,luo2025sonic}. However, these methods mainly emphasize feasible motion generation and kinematic tracking, whereas ThorArena evaluates policies using paired motion--force demonstrations and replayed real-world interaction-force disturbances.

\subsection{Humanoid Motion Datasets and Evaluation}

Human motion datasets and body representations provide large-scale motion
capture data and unified pose formats for motion modeling and policy
training \cite{loper2023smpl,AMASS:2019,harvey2020robust}. However, these
resources are primarily designed for human motion modeling or animation and
do not explicitly account for humanoid embodiment constraints or physical
interaction.

Recent humanoid-oriented datasets convert human motions, Internet videos,
or teleoperated demonstrations into robot-compatible trajectories. Some
works emphasize large-scale data collection and physically plausible
retargeting \cite{mao2024learning,lee2025phuma}, while others extend the
scope to human--object, humanoid manipulation, and humanoid--scene
interaction \cite{Lu_2025_HUMOTO,zhao2025humanoid,liu2024mimicking}.
Existing benchmarks commonly evaluate tracking error, task success, motion difficulty, or perceived human-likeness~\cite{meng2025benchmarking,li2026omniclone,li2026towards}.
Despite this progress, measured interaction forces are rarely incorporated
into standardized policy evaluation. ThorArena addresses this gap by
combining synchronized motion--force demonstrations with force-replay-based
evaluation.

\section{METHOD}
In this section, we present the overall methodology for constructing a force-aware humanoid motion benchmark. The proposed framework consists of three components. First, we collect real-world force-interaction demonstrations that pair human motion with synchronized two-hand force measurements. Second, we define force-aware evaluation metrics to assess whole-body tracking accuracy, robustness under different contact-force levels, control effort, and episode survival. Third, we develop a benchmark evaluation protocol that replays recorded interaction forces during policy rollout and provides a unified policy-adapter interface for evaluating different humanoid motion policies under the same force-replay protocol. Together, these components enable systematic evaluation of humanoid motion policies in contact-rich scenarios.

\subsection{Real-World Force-Interaction Data Acquisition} 
\label{sec:data_acquisition}

\begin{figure}[h]
    \centering
    \includegraphics[
        width=1\linewidth,
        trim=4pt 4pt 4pt 4pt,
        clip
    ]{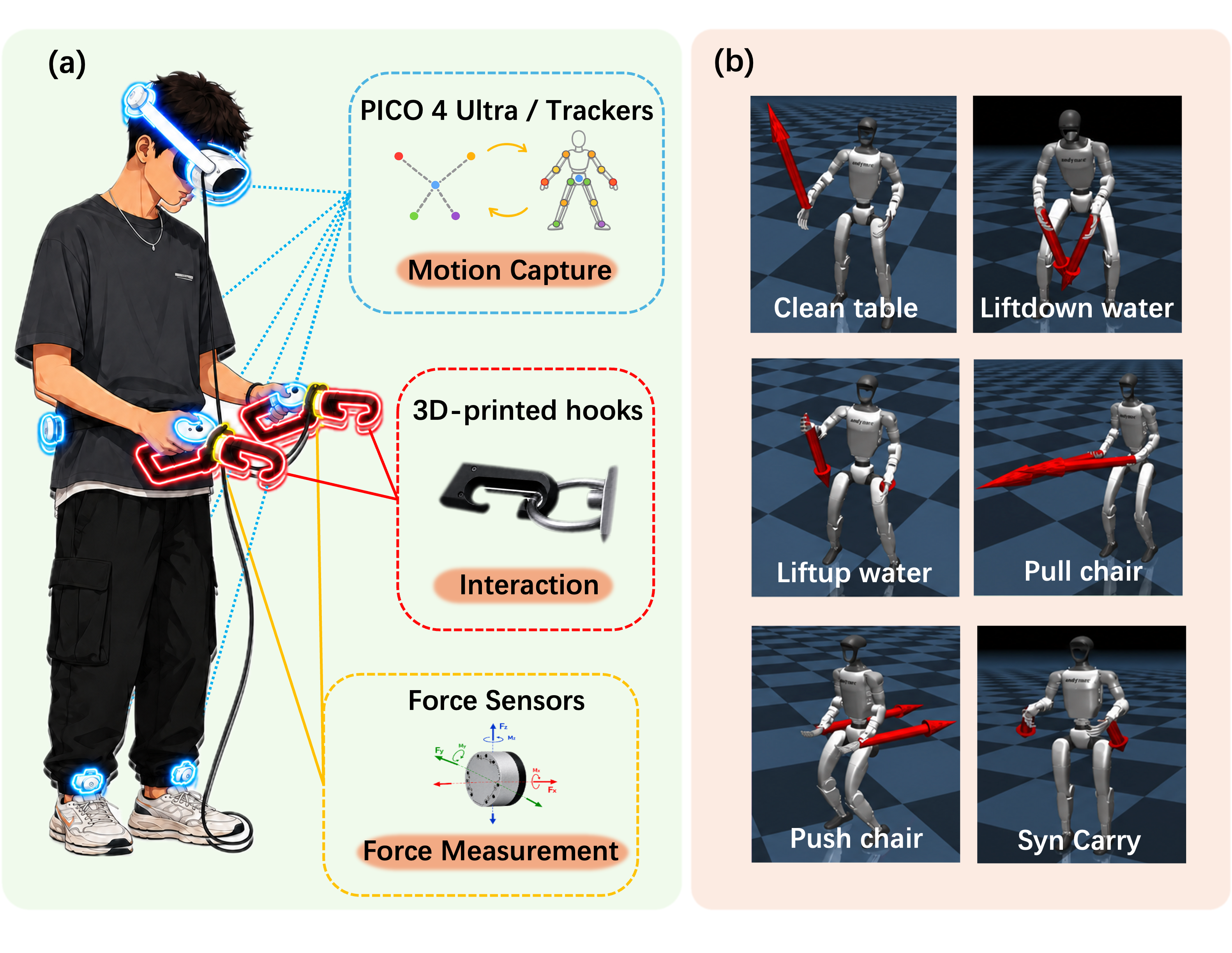}
    \caption{\textbf{Real-world motion--force acquisition system and task set.}
    (a) The operator uses a VR-based system for whole-body motion capture, while two force sensors coupled with 3D-printed hooks record the forces exerted through both hands and provide stable attachment points for object interaction.
    (b) Retargeted humanoid motions for the six force-interaction tasks:
    \textbf{Clean Table} (table wiping), \textbf{Liftdown Water} (lowering and placing a water container), \textbf{Liftup Water} (lifting a water container), \textbf{Pull Chair} (pulling heavy objects), \textbf{Push Chair} (pushing heavy objects), and \textbf{Syn Carry} (cooperative carrying).
    Red arrows indicate the dominant interaction-force directions.}
    \label{fig:data_acquisition_tasks}
\end{figure}

To collect synchronized human motion and interaction-force demonstrations,
we build the real-world acquisition system shown in
Fig.~\ref{fig:data_acquisition_tasks}(a). The operator wears a PICO 4 Ultra
headset and body trackers to capture whole-body motion. Two force sensors
coupled with 3D-printed hooks measure the forces exerted through the left
and right hands, while the hooks provide stable attachment points for
interaction with external objects.

We design six representative real-world force-interaction tasks to cover
common physical interaction patterns. As illustrated in
Fig.~\ref{fig:data_acquisition_tasks}(b), the task set includes table
wiping, object lowering, object lifting, chair pulling, chair pushing, and
cooperative carrying. Compared with conventional human motion data, the
collected demonstrations contain not only body poses and motion
trajectories, but also synchronized force measurements generated during
physical interaction. The resulting paired motion--force demonstrations
support subsequent motion retargeting, benchmark construction, and
force-aware humanoid policy evaluation.

\begin{figure*}[!t]
\centering
\includegraphics[width=0.95\textwidth]{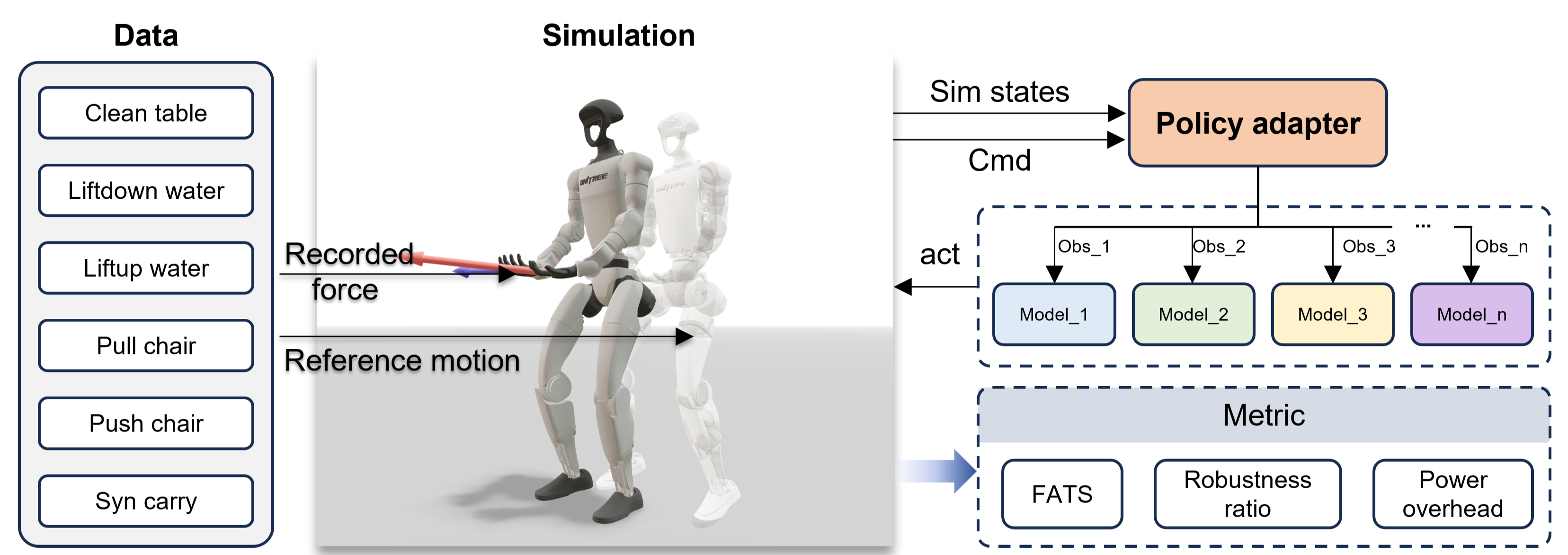}
\caption{\textbf{ThorArena benchmark evaluation pipeline.}
The benchmark replays retargeted reference motions and recorded two-hand forces in simulation, evaluates different whole-body control policies through a unified policy adapter, and reports FATS and complementary diagnostic metrics under force and no-force settings.}
\label{fig:evaluation}
\end{figure*}

In terms of data scale, we collect 60 raw demonstration sequences for each task, resulting in 360 raw sequences in total. Each sequence corresponds to a complete task execution and captures the motion-force evolution throughout the interaction process. Across demonstrations, the measured forces vary in both magnitude and direction, which increases the diversity of force-interaction patterns and provides representative data for evaluating humanoid behaviors in contact-rich scenarios.

The collected raw demonstrations are stored as independent recording sequences. Each sequence contains time-aligned human body keypoint motion and left/right three-axis force measurements. The force measurements are acquired from the sensors mounted on both sides and represented as three-dimensional force vectors. This format preserves the original correspondence between motion and force during data collection, providing a basis for subsequent data processing, motion retargeting, benchmark construction, and force-aware policy evaluation.

\subsection{Force-Aware Evaluation Metrics}
\label{sec:force_metrics}

To evaluate whole-body motion tracking under interaction forces, we adopt metrics that jointly characterize tracking accuracy, force robustness, control effort, and episode survival. The primary metric is the Force-Aware Tracking Score (FATS).

The basic tracking error is the whole-body keypoint error in the world frame:
\begin{equation}
e_{\mathrm{kp}}(t)
=
\frac{1}{N}
\sum_{k=1}^{N}
\bigl\|
\mathbf{x}^{\mathrm{ref}}_{k,t} - \mathbf{x}_{k,t}
\bigr\|_2 ,
\label{eq:kp_err}
\end{equation}
where $N$ is the number of body keypoints, and $\mathbf{x}^{\mathrm{ref}}_{k,t}$ and $\mathbf{x}_{k,t}$ denote the reference and measured positions of keypoint $k$ at timestep $t$. The reference keypoints are re-anchored to the robot's current root pose at every timestep, so the error measures root-relative pose fidelity and is invariant to global root drift, which is reported separately as a diagnostic.

Timesteps are stratified into low-, medium-, and high-force regimes by the total applied hand force, using its 33rd and 66th percentiles within each dataset as thresholds. For each episode $i$, the keypoint RMSE in the three regimes, $E_i^{\mathrm{low}}$, $E_i^{\mathrm{mid}}$, and $E_i^{\mathrm{high}}$, is aggregated into a force-aware tracking error
\begin{equation}
E_i
=
\sqrt{
w_{\mathrm{low}} \bigl(E_i^{\mathrm{low}}\bigr)^{2}
+
w_{\mathrm{mid}} \bigl(E_i^{\mathrm{mid}}\bigr)^{2}
+
w_{\mathrm{high}} \bigl(E_i^{\mathrm{high}}\bigr)^{2}
} ,
\label{eq:weighted_err}
\end{equation}
where $(w_{\mathrm{low}}, w_{\mathrm{mid}}, w_{\mathrm{high}}) = (0.2, 0.3, 0.5)$ emphasizes the high-force regime.

The episode-level score combines the tracking error with a survival factor $s_i = \min\bigl(T_i / T_i^{\mathrm{ref}},\, 1\bigr)$, where $T_i$ is the achieved episode length and $T_i^{\mathrm{ref}}$ the reference horizon:
\begin{equation}
S_i
=
100
\exp\!\left(-\frac{E_i}{\sigma}\right)
s_i ,
\qquad
\mathrm{FATS}
=
\frac{1}{M}
\sum_{i=1}^{M} S_i ,
\label{eq:fats}
\end{equation}
where $\sigma = 0.15\,\mathrm{m}$ controls the error sensitivity and $M$ is the number of episodes. In the no-force setting, the same formulation is applied with force replay disabled.

We additionally report diagnostic metrics, including the robustness ratio $\rho$ and the power overhead $\eta$:
\begin{equation}
\rho
=
\frac{E^{\mathrm{low}}}{E^{\mathrm{high}}} ,
\qquad
\eta
=
\frac{P^{\mathrm{high}}}{P^{\mathrm{low}}} ,
\label{eq:diagnostics}
\end{equation}
where $E^{\mathrm{low}}$, $E^{\mathrm{high}}$ and $P^{\mathrm{low}}$, $P^{\mathrm{high}}$ denote the keypoint RMSE and average mechanical power in the low- and high-force regimes. A robustness ratio closer to one indicates less degradation under stronger forces, while a lower power overhead indicates less additional control effort. These diagnostics do not enter the FATS formula.

\begin{figure*}[!t]
\centering
\includegraphics[width=0.95\textwidth]{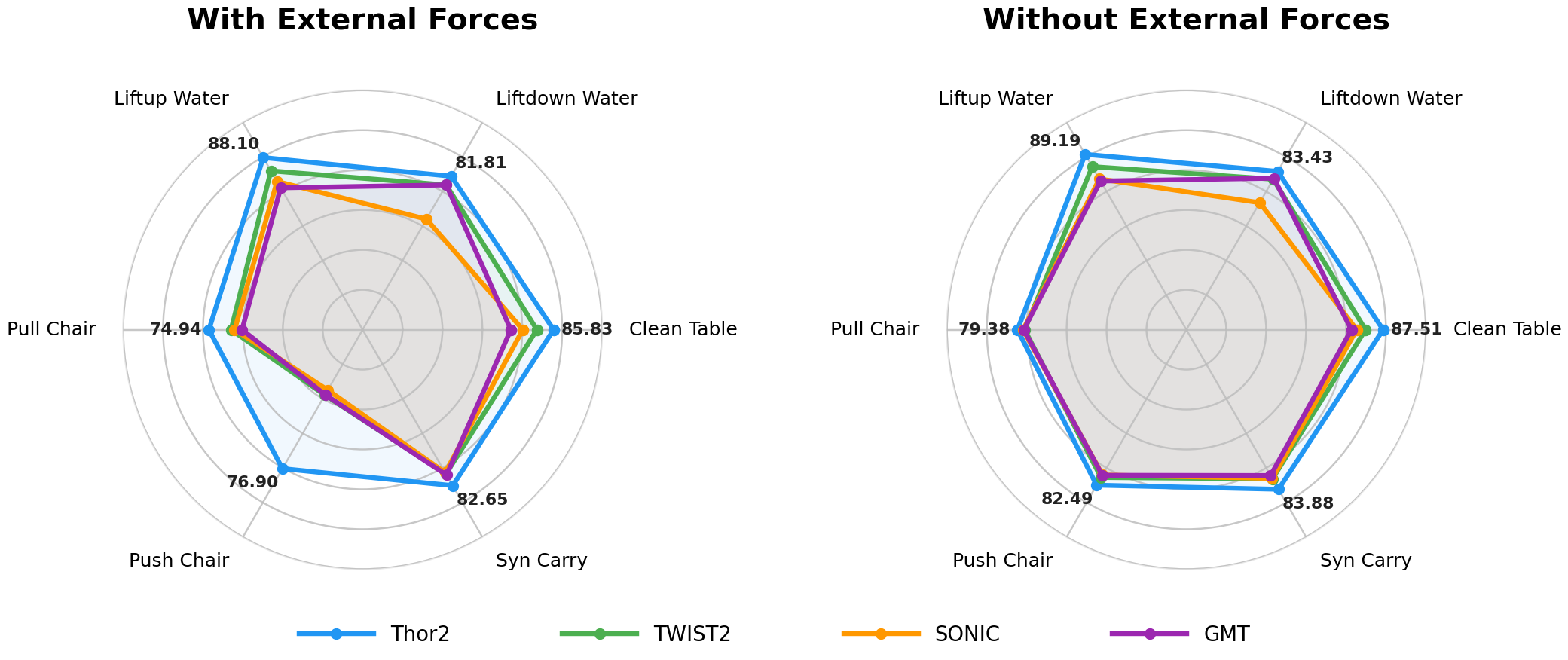}
\caption{Task-level FATS comparison with and without external forces.}
\label{fig:fats_radar}
\end{figure*}

\subsection{Benchmark Evaluation Protocol}
\label{sec:benchmark_protocol}

As illustrated in Fig.~\ref{fig:evaluation}, the benchmark integrates
retargeted reference motions, synchronized two-hand force recordings,
physics-based simulation, and different whole-body control policies through
a unified policy-adapter interface. Given a reference motion and its
corresponding force recordings, the benchmark runner synchronously replays
the recorded forces during policy rollout, advances the simulator, handles
episode resets, and records rollout traces. Policy-specific observation
construction, inference, and action conversion are encapsulated in
lightweight adapters, allowing different policy architectures to be
evaluated under the same benchmark runner and force-replay protocol.

In the force-replay protocol, the benchmark loads a specified evaluation dataset and replays the recorded left and right force sequences synchronously with the reference motion. The recorded sensor-frame force vectors are first transformed into the hand-local frame and then converted to the world frame according to the simulated hand poses. The transformed forces are applied to the left and right hand bodies during simulation. A scalar force coefficient controls the applied force magnitude: it is set to (1) for external-force evaluation, reproducing the measured interaction forces, and set to (0) for no-force evaluation, disabling external force application.

For the official benchmark evaluation, we use a full-coverage protocol over the evaluation motion set. The benchmark constructs a deterministic schedule over all motions in the evaluation dataset, ensuring that every demonstration sequence is evaluated once under the same force-replay protocol. Multiple parallel simulation environments are used to improve evaluation efficiency while preserving deterministic coverage of the motion set.

To support different humanoid motion policies, the benchmark uses a unified policy-adapter interface. Each adapter receives the current simulator state and reference-motion command, constructs the policy-specific observation from available benchmark signals, performs policy inference, and converts the predicted action into the benchmark action representation. Depending on the input requirements of the target policy, the adapter may use reference motion features, robot proprioception, action histories, or future motion information. This design separates the benchmark mechanics from model-specific observation layouts, action parameterizations, and joint-order conventions, allowing different policy architectures to be evaluated under the same force-replay and benchmark protocol without changing the core evaluation pipeline.

ThorArena evaluates each policy as a complete control system under a fixed
deployment-compatible dynamics profile derived from its original training
or deployment configuration. Each profile specifies the robot model,
control and physics timesteps, action scaling, and action delay required by
the corresponding policy interface. The profiles are fixed before
evaluation and remain unchanged across all tasks and demonstration
sequences. Across methods, the reference data, force-replay procedure,
episode handling, metric computation, and rollout-recording protocol are
held fixed.

\section{EXPERIMENTS}
\label{sec:experiments}

We evaluate ThorArena on the collected humanoid force-interaction dataset, which consists of six subtasks: \textit{clean\_table}, \textit{liftdown\_water}, \textit{liftup\_water}, \textit{pull\_chair}, \textit{push\_chair}, and \textit{syn\_carry}. Four whole-body control policies are compared: Thor2, TWIST2~\cite{ze2025twist2}, GMT~\cite{chen2025gmt}, and SONIC~\cite{luo2025sonic}. For each policy, we conduct evaluations under two settings: with recorded external interaction forces and without external forces. We use FATS as the primary evaluation metric and further analyze diagnostic metrics including survival rate, body tracking error, force robustness, and control effort. Unless otherwise specified, aggregate metrics are averaged over the six subtasks.

Fig.~\ref{fig:fats_radar} provides an overview of the task-level FATS distribution under both force and no-force conditions. Thor2 consistently occupies the outer region of the radar plots, indicating the best overall tracking performance across the six subtasks. The comparison also shows that external forces cause more pronounced performance degradation for the other policies, especially on \textit{push\_chair}.

\subsection{Experiments with External Forces}

The external-force setting evaluates whether a humanoid control policy can maintain stable and accurate motion tracking during physical interaction. Table~\ref{tab:force_task_fats} reports task-level FATS results, and Table~\ref{tab:force_diagnostics} summarizes the diagnostic metrics averaged over tasks. Thor2 achieves the best overall performance, ranking first on all six subtasks and obtaining the highest average FATS of 81.71 with a near-perfect survival rate.

\begin{table}[!h]
\caption{Task-level FATS under external forces.}
\label{tab:force_task_fats}
\centering
\scriptsize
\renewcommand{\arraystretch}{1.25}
\setlength{\tabcolsep}{3.5pt}
\begin{tabular}{lcccc}
\noalign{\hrule height 0.9pt}
Task & Thor2 & TWIST2~\cite{ze2025twist2} & GMT~\cite{chen2025gmt} & SONIC~\cite{luo2025sonic} \\
\hline
\textit{clean\_table} & $\textbf{85.83}$ & 80.94 & 73.33 & 76.82 \\
\textit{liftdown\_water} & $\textbf{81.81}$ & 78.86 & 78.88 & 67.31 \\
\textit{liftup\_water} & $\textbf{88.10}$ & 83.63 & 77.90 & 80.05 \\
\textit{pull\_chair} & $\textbf{74.94}$ & 68.46 & 65.26 & 67.42 \\
\textit{push\_chair} & $\textbf{76.90}$ & 52.07 & 51.93 & 50.37 \\
\textit{syn\_carry} & $\textbf{82.65}$ & 78.71 & 79.05 & 78.24 \\
\noalign{\hrule height 0.9pt}
\end{tabular}
\renewcommand{\arraystretch}{1.0}
\end{table}

\begin{table}[h]
\caption{Diagnostic metrics averaged over tasks under external forces.}
\label{tab:force_diagnostics}
\centering
\scriptsize
\renewcommand{\arraystretch}{1.25}
\setlength{\tabcolsep}{3.5pt}
\begin{tabular}{lcccc}
\noalign{\hrule height 0.9pt}
Metric & Thor2 & \mbox{TWIST2~\cite{ze2025twist2}} & \mbox{GMT~\cite{chen2025gmt}} & \mbox{SONIC~\cite{luo2025sonic}} \\
\hline
FATS & $\textbf{81.71}$ & 73.78 & 71.06 & 70.04 \\
Survival & $\textbf{1.000}$ & 0.941 & 0.916 & 0.957 \\
Low KP & $\textbf{27.8}$ & 37.5 & 37.8 & 43.9 \\
High KP & $\textbf{33.8}$ & 40.7 & 42.4 & 54.1 \\
Robustness & 0.812 & $\textbf{0.908}$ & 0.888 & 0.816 \\
Power & 1.27$\times$ & 1.23$\times$ & 1.37$\times$ & $\textbf{1.22}\times$ \\
Upper KP & $\textbf{27.5}$ & 35.7 & 38.8 & 55.4 \\
Lower KP & $\textbf{42.8}$ & 48.5 & 48.4 & 54.4 \\
FATS Std & $\textbf{2.96}$ & 9.57 & 12.75 & 9.02 \\
\noalign{\hrule height 0.9pt}
\end{tabular}
\renewcommand{\arraystretch}{1.0}
\end{table}

The other three policies obtain relatively close average FATS scores but exhibit distinct performance profiles. TWIST2 attains the highest robustness ratio $\rho = \mathrm{RMSE}_{\mathrm{low}} / \mathrm{RMSE}_{\mathrm{high}}$, defined as the ratio of keypoint RMSE in low-force to high-force segments ($\rho \to 1$ indicates slower degradation); GMT tracks more accurately than SONIC under strong interaction forces; and SONIC favors episode stability over tracking precision. We examine these trade-offs in detail below.

\begin{figure}[h]
\centering
\includegraphics[width=1\linewidth]{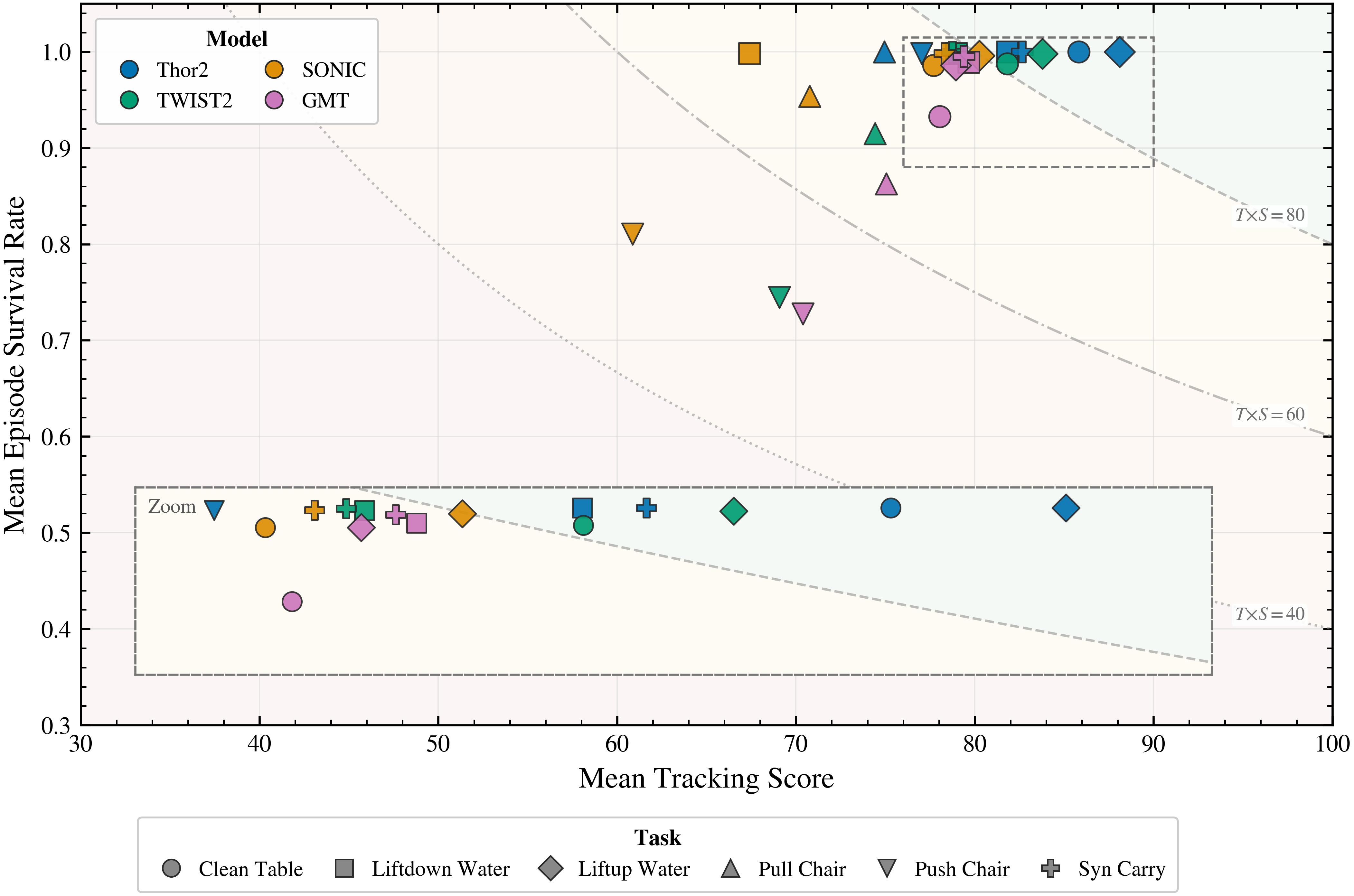}
\caption{Tracking score and survival distribution under external forces.}
\label{fig:tracking_survival_scatter}
\end{figure}

Fig.~\ref{fig:tracking_survival_scatter} visualizes the relationship between tracking score and episode survival under external forces. Thor2 maintains both high tracking scores and nearly full survival ($\geq 0.99$ on all tasks), while the other policies exhibit substantially lower mean episode survival during force interaction. Among all subtasks, \textit{push\_chair} causes the largest degradation for TWIST2, GMT, and SONIC, whose survival drops to 0.73--0.81. Unlike the water-carrying tasks, where vertical loads mainly perturb upper-body tracking, pushing imposes sustained horizontal forces that directly threaten balance, making it the most challenging interaction in the benchmark.

\begin{figure}[h]
\centering
\includegraphics[width=1\linewidth]{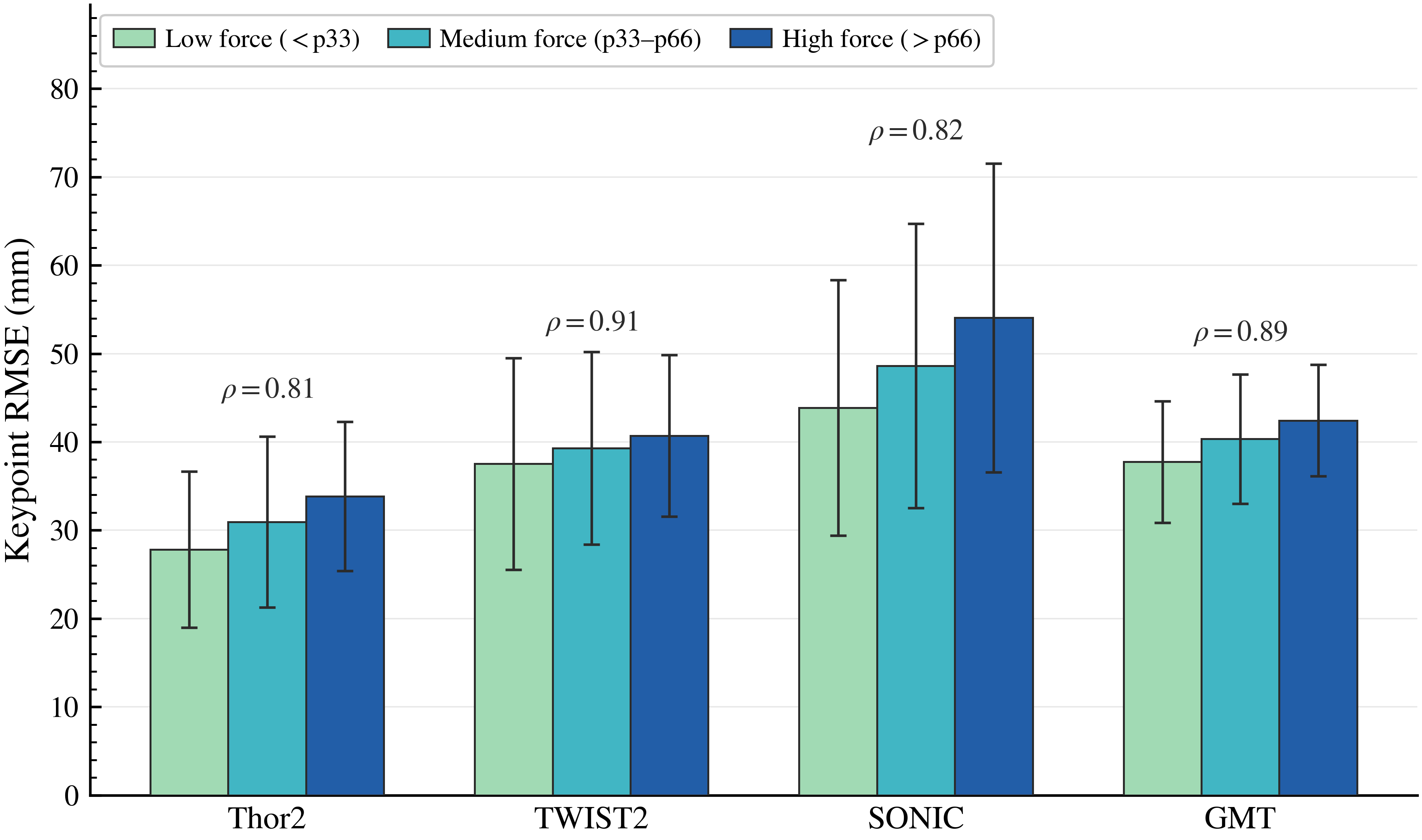}
\caption{Keypoint tracking errors across low-force and high-force segments.}
\label{fig:force_buckets}
\end{figure}

We further analyze performance across force intensities in Fig.~\ref{fig:force_buckets}. Thor2 achieves the lowest absolute keypoint error in both force regimes (27.8\,mm / 33.8\,mm, about 25\% lower than the second best), maintaining superior tracking accuracy under external forces. Although TWIST2 attains the highest $\rho$, we note that $\rho$ is a relative measure: its high ratio partly reflects a larger low-force baseline error, and should therefore be interpreted jointly with the absolute errors in Table~\ref{tab:force_diagnostics}. SONIC exhibits the opposite trade-off: it attains the lowest power overhead and the second-highest survival rate---notably outperforming TWIST2 and GMT on the two most demanding tasks---yet incurs the largest keypoint errors, concentrated in upper-body segments under sustained force. This suggests that SONIC responds to perturbations compliantly, preserving balance rather than actively rejecting disturbances, at the cost of force-aware tracking precision.

\subsection{Experiments without External Forces}

The no-force setting evaluates the nominal motion-tracking ability of each policy without physical perturbation. Table~\ref{tab:noforce_results} shows that all policies achieve high survival rates close to 1.0, indicating that the performance differences mainly come from tracking accuracy rather than falling or early termination. Thor2 still performs best, achieving a FATS of 84.31 and the lowest keypoint error of 26.1 mm. TWIST2 ranks second, followed by GMT and SONIC.

\begin{table}[h]
\caption{Performance comparison under no-force evaluation.}
\label{tab:noforce_results}
\centering
\scriptsize
\renewcommand{\arraystretch}{1.25}
\setlength{\tabcolsep}{2.7pt}
\begin{tabular}{lccccc}
\noalign{\hrule height 0.9pt}
Model & FATS & Survival & KP Error & Upper KP & Lower KP \\
\hline
Thor2 
& $\textbf{84.31}$ & $\textbf{1.000}$ & $\textbf{26.1}$ & $\textbf{15.3}$ & 39.9 \\
TWIST2~\cite{ze2025twist2}
& 80.96 & 0.997 & 32.0 & 26.6 & 39.8 \\
GMT~\cite{chen2025gmt} 
& 79.25 & 0.992 & 34.5 & 30.9 & 40.0 \\
SONIC~\cite{luo2025sonic}
& 78.43 & 0.997 & 36.2 & 35.3 & $\textbf{39.1}$ \\
\noalign{\hrule height 0.9pt}
\end{tabular}
\renewcommand{\arraystretch}{1.0}
\end{table}

Compared with the external-force setting, the no-force results show smaller performance gaps among policies. The survival rates are almost saturated, and the lower-body errors are also close across different policies. Therefore, the remaining performance differences mainly come from upper-body tracking accuracy. This contrast shows that conventional no-force evaluation is insufficient for contact-rich humanoid tasks: it can measure nominal tracking quality, but it cannot reveal force-induced instability, robustness degradation, or additional control effort. ThorArena therefore provides a more complete evaluation by comparing both force and no-force conditions.

\section{CONCLUSIONS}

In this paper, we presented ThorArena, a force-aware benchmark for evaluating
humanoid motion policies under recorded interaction-force disturbances.
ThorArena combines synchronized human motion--force demonstrations, a
simulation-based force-replay protocol, and force-aware evaluation metrics,
including FATS. Experiments on representative whole-body control policies
under matched force and no-force settings show that ThorArena reveals
differences in force robustness, control effort, and episode survival that
are difficult to capture through conventional no-force evaluation.

ThorArena provides a practical and reproducible benchmark for studying
force-aware humanoid interaction. Future work will expand the dataset with
more diverse tasks, objects, and force conditions, while improving the data
acquisition pipeline in terms of sensing accuracy, synchronization, and
collection scalability. We will further validate the force-level
stratification and metric formulation through systematic experiments,
extend the benchmark to additional humanoid embodiments and control systems,
and conduct physical-robot evaluations to assess its real-world
applicability and consistency with simulation-based results.










\balance
\bibliographystyle{IEEEtran}
\bibliography{wbc,benchmark_dataset}

@inproceedings{AMASS:2019,
  title={AMASS: Archive of Motion Capture as Surface Shapes},
  author={Mahmood, Naureen and Ghorbani, Nima and F. Troje, Nikolaus and Pons-Moll, Gerard and Black, Michael J.},
  booktitle = {The IEEE International Conference on Computer Vision (ICCV)},
  year={2019},
  month = {Oct},
  url = {https://amass.is.tue.mpg.de},
  month_numeric = {10}
}

@article{mao2024learning,
  title={Learning from massive human videos for universal humanoid pose control},
  author={Mao, Jiageng and Zhao, Siheng and Song, Siqi and Shi, Tianheng and Ye, Junjie and Zhang, Mingtong and Geng, Haoran and Malik, Jitendra and Guizilini, Vitor and Wang, Yue},
  journal={arXiv preprint arXiv:2412.14172},
  year={2024}
}

@article{zhao2025humanoid,
  title={Humanoid everyday: A comprehensive robotic dataset for open-world humanoid manipulation},
  author={Zhao, Zhenyu and Jing, Hongyi and Liu, Xiawei and Mao, Jiageng and Jha, Abha and Yang, Hanwen and Xue, Rong and Zakharor, Sergey and Guizilini, Vitor and Wang, Yue},
  journal={arXiv preprint arXiv:2510.08807},
  year={2025}
}

@InProceedings{Lu_2025_HUMOTO,
      author    = {Lu, Jiaxin and Huang, Chun-Hao Paul and Bhattacharya, Uttaran and Huang, Qixing and Zhou, Yi},
      title     = {HUMOTO: A 4D Dataset of Mocap Human Object Interactions},
      booktitle = {Proceedings of the IEEE/CVF International Conference on Computer Vision (ICCV)},
      month     = {October},
      year      = {2025},
      pages     = {10886-10897}
}

@article{harvey2020robust,
  title={Robust motion in-betweening},
  author={Harvey, F{\'e}lix G and Yurick, Mike and Nowrouzezahrai, Derek and Pal, Christopher},
  journal={ACM Transactions on Graphics (TOG)},
  volume={39},
  number={4},
  pages={60--1},
  year={2020},
  publisher={ACM New York, NY, USA}
}

@inproceedings{Luo2023PerpetualHC,
    author={Zhengyi Luo and Jinkun Cao and Alexander W. Winkler and Kris Kitani and Weipeng Xu},
    title={Perpetual Humanoid Control for Real-time Simulated Avatars},
    booktitle={International Conference on Computer Vision (ICCV)},
    year={2023}
}

@article{lee2025phuma,
  title={PHUMA: Physically Reliable Humanoid Locomotion Dataset},
  author={Kyungmin Lee and Sibeen Kim and Youngdo Lee and Minho Park and Hyunseung Kim and Dongyoon Hwang and Donghu Kim and Hojoon Lee and Jaegul Choo},
  journal={arXiv preprint arXiv:2510.26236},
  year={2025},
}

@incollection{loper2023smpl,
  title={SMPL: A skinned multi-person linear model},
  author={Loper, Matthew and Mahmood, Naureen and Romero, Javier and Pons-Moll, Gerard and Black, Michael J},
  booktitle={Seminal Graphics Papers: Pushing the Boundaries, Volume 2},
  pages={851--866},
  year={2023}
}

@article{meng2025benchmarking,
  title={Benchmarking Humanoid Imitation Learning with Motion Difficulty},
  author={Meng, Zhaorui and Yin, Lu and Chen, Xinrui and Chen, Anjun and Guo, Shihui and Qin, Yipeng},
  journal={arXiv preprint arXiv:2512.07248},
  year={2025}
}

@article{liu2024mimicking,
  title={Mimicking-bench: A benchmark for generalizable humanoid-scene interaction learning via human mimicking},
  author={Liu, Yun and Yang, Bowen and Zhong, Licheng and Wang, He and Yi, Li},
  journal={arXiv preprint arXiv:2412.17730},
  year={2024}
}

@article{li2026omniclone,
  title={OmniClone: Engineering a Robust, All-Rounder Whole-Body Humanoid Teleoperation System},
  author={Li, Yixuan and Ma, Le and Lin, Yutang and Du, Yushi and Liu, Mengya and Hu, Kaizhe and Cui, Jieming and Zhu, Yixin and Liang, Wei and Jia, Baoxiong and Huang, Siyuan},
  journal={arXiv preprint arXiv:2603.14327},
  year={2026}
}

@inproceedings{li2026towards,
  title={Towards motion turing test: Evaluating human-likeness in humanoid robots},
  author={Li, Mingzhe and Liu, Mengyin and Wu, Zekai and Lin, Xincheng and Zhang, Junsheng and Yan, Ming and Xie, Zengye and Zhang, Changwang and Wen, Chenglu and Xu, Lan and others},
  booktitle={Proceedings of the IEEE/CVF Conference on Computer Vision and Pattern Recognition},
  pages={16486--16498},
  year={2026}
}

@article{yang2025omniretarget,
  title={Omniretarget: Interaction-preserving data generation for humanoid whole-body loco-manipulation and scene interaction},
  author={Yang, Lujie and Huang, Xiaoyu and Wu, Zhen and Kanazawa, Angjoo and Abbeel, Pieter and Sferrazza, Carmelo and Liu, C Karen and Duan, Rocky and Shi, Guanya},
  journal={arXiv preprint arXiv:2509.26633},
  year={2025}
}

@article{li2025amo,
title={AMO: Adaptive Motion Optimization for Hyper-Dexterous Humanoid Whole-Body Control},
author={Li, Jialong and Cheng, Xuxin and Huang, Tianshu and Yang, Shiqi and Qiu, Rizhao and Wang, Xiaolong},
journal={Robotics: Science and Systems 2025},
year={2025}
}

@misc{li2025clone,
  title={CLONE: Closed-Loop Whole-Body Humanoid Teleoperation for Long-Horizon Tasks}, 
  author={Yixuan Li and Yutang Lin and Jieming Cui and Tengyu Liu and Wei Liang and Yixin Zhu and Siyuan Huang},
  journal={arXiv preprint arXiv:2506.08931}, 
  year={2025}
}

@article{zhu2026clot,
  title={CLOT: Closed-Loop Global Motion Tracking for Whole-Body Humanoid Teleoperation},
  author={Zhu, Tengjie and Cai, Guanyu and Zhaohui, Yang and Ren, Guanzhu and Xie, Haohui and Wang, ZiRui and Wu, Junsong and Wang, Jingbo and Yang, Xiaokang and Mu, Yao and others},
  journal={arXiv preprint arXiv:2602.15060},
  year={2026}
}

@article{joao2025gmr,
  title={Retargeting Matters: General Motion Retargeting for Humanoid Motion Tracking},
  author= {Joao Pedro Araujo and Yanjie Ze and Pei Xu and Jiajun Wu and C. Karen Liu},
  year= {2025},
  journal= {arXiv preprint arXiv:2510.02252}
}

@article{chen2025gmt,
  title={Gmt: General motion tracking for humanoid whole-body control},
  author={Chen, Zixuan and Ji, Mazeyu and Cheng, Xuxin and Peng, Xuanbin and Peng, Xue Bin and Wang, Xiaolong},
  journal={arXiv preprint arXiv:2506.14770},
  year={2025}
}

@inproceedings{he2024learning,
  title={Learning human-to-humanoid real-time whole-body teleoperation},
  author={He, Tairan and Luo, Zhengyi and Xiao, Wenli and Zhang, Chong and Kitani, Kris and Liu, Changliu and Shi, Guanya},
  booktitle={2024 IEEE/RSJ International Conference on Intelligent Robots and Systems (IROS)},
  pages={8944--8951},
  year={2024},
  organization={IEEE}
}

@article{ben2025homie,
  title={Homie: Humanoid loco-manipulation with isomorphic exoskeleton cockpit},
  author={Ben, Qingwei and Jia, Feiyu and Zeng, Jia and Dong, Junting and Lin, Dahua and Pang, Jiangmiao},
  journal={arXiv preprint arXiv:2502.13013},
  year={2025}
}

@inproceedings{he2025hover,
  title={Hover: Versatile neural whole-body controller for humanoid robots},
  author={He, Tairan and Xiao, Wenli and Lin, Toru and Luo, Zhengyi and Xu, Zhenjia and Jiang, Zhenyu and Kautz, Jan and Liu, Changliu and Shi, Guanya and Wang, Xiaolong and others},
  booktitle={2025 IEEE International Conference on Robotics and Automation (ICRA)},
  pages={9989--9996},
  year={2025},
  organization={IEEE}
}

@article{fu2024humanplus,
  title={Humanplus: Humanoid shadowing and imitation from humans},
  author={Fu, Zipeng and Zhao, Qingqing and Wu, Qi and Wetzstein, Gordon and Finn, Chelsea},
  journal={arXiv preprint arXiv:2406.10454},
  year={2024}
}

@article{he2024omnih2o,
  title={Omnih2o: Universal and dexterous human-to-humanoid whole-body teleoperation and learning},
  author={He, Tairan and Luo, Zhengyi and He, Xialin and Xiao, Wenli and Zhang, Chong and Zhang, Weinan and Kitani, Kris and Liu, Changliu and Shi, Guanya},
  journal={arXiv preprint arXiv:2406.08858},
  year={2024}
}

@article{luo2025sonic,
    title={SONIC: Supersizing Motion Tracking for Natural Humanoid Whole-Body Control},
    author={Luo, Zhengyi and Yuan, Ye and Wang, Tingwu and Li, Chenran and Chen, Sirui and Casta\~neda, Fernando and Cao, Zi-Ang and Li, Jiefeng and Minor, David and Ben, Qingwei and Da, Xingye and Ding, Runyu and Hogg, Cyrus and Song, Lina and Lim, Edy and Jeong, Eugene and He, Tairan and Xue, Haoru and Xiao, Wenli and Wang, Zi and Yuen, Simon and Kautz, Jan and Chang, Yan and Iqbal, Umar and Fan, Linxi and Zhu, Yuke},
    journal={arXiv preprint arXiv:2511.07820},
    year={2025}
}

@article{ze2025twist,
title={TWIST: Teleoperated Whole-Body Imitation System},
author= {Yanjie Ze and Zixuan Chen and João Pedro Araújo and Zi-ang Cao and Xue Bin Peng and Jiajun Wu and C. Karen Liu},
year= {2025},
journal= {arXiv preprint arXiv:2505.02833}
}

@article{ze2025twist2,
title={TWIST2: Scalable, Portable, and Holistic Humanoid Data Collection System},
author= {Yanjie Ze and Siheng Zhao and Weizhuo Wang and Angjoo Kanazawa and Rocky Duan and Pieter Abbeel and Guanya Shi and Jiajun Wu and C. Karen Liu},
year= {2025},
journal= {arXiv preprint arXiv:2511.02832}
}

\end{document}